\documentclass[letter, 10 pt, conference]{ieeeconf}  

\IEEEoverridecommandlockouts 
\usepackage[english]{babel}
\usepackage[utf8]{inputenc}
\usepackage{times} 
\usepackage{cite}

\DeclareMathAlphabet{\pazocal}{OMS}{zplm}{m}{n}

\usepackage{hyperref}
\usepackage[nolist]{acronym}
\usepackage{balance} 

\usepackage{multirow}

\usepackage{siunitx}
\usepackage{graphicx} 
\usepackage{epsfig} 
\usepackage{pgfplots}
\usepackage{caption}
\usepackage[font=footnotesize]{subcaption}
\usepackage{pgfplots}

\usepackage{float}

\usepackage[export]{adjustbox}

\usepackage{xcolor}
\definecolor{myYellow}{rgb}{0.93,0.69,0.13}
\definecolor{myPurple}{rgb}{0.49,0.18,0.56}
\definecolor{myGreen}{rgb}{0.26 0.72 0.54}

\usepackage{mathtools}
\usepackage{nicefrac}
\usepackage{amsmath}
\usepackage{amssymb}
\usepackage{bm} 

\DeclareMathOperator*{\minimize}{minimize}
\DeclareMathOperator*{\maximize}{maximize}

\usepackage{etoolbox}
\makeatletter%
\AfterPreamble{%
	\usepackage{hyperref}%
	\let\oldhypertarget\hypertarget%
	\renewcommand{\hypertarget}[2]{%
		\oldhypertarget{#1}{#2}%
		\protected@write\@mainaux{}{%
			\string\expandafter\string\gdef%
			\string\csname\string\detokenize{#1}\string\endcsname{#2}%
		}%
	}%
	\newcommand{\myhyperlink}[1]{%
		\hyperlink{#1}{\csname #1\endcsname}%
	}%
}
\makeatother%

\newcounter{Remark}
\newcounter{Definition}
\newcounter{Problem}
\usepackage{algorithm}
\usepackage[noend]{algpseudocode}

\makeatletter
\def\BState{\State\hskip-\ALG@thistlm}
\makeatother

\usepackage{tikz}
\pgfdeclarelayer{foreground}
\pgfsetlayers{background, main, foreground}
\usetikzlibrary{quotes, angles, backgrounds, arrows, automata, shapes, positioning, calc, through, spy, decorations.pathreplacing, decorations.markings, arrows.meta, automata, petri}

\tikzset{
    imglabel/.style={
      rectangle,
      inner sep=2pt,
      text=black,
      minimum height=1em,
      text centered,
      fill=white,
      fill opacity=1.0,
      text opacity=1,
      anchor=south west,
    },
  }
\tikzset{
	state/.style={
		rectangle,
		draw=black, very thick,
		minimum height=1.0em,
		text centered,
	},
}
\tikzset{
  on each segment/.style={
    decorate,
    decoration={
      show path construction,
      moveto code={},
      lineto code={
        \path [#1]
        (\tikzinputsegmentfirst) -- (\tikzinputsegmentlast);
      },
      curveto code={
        \path [#1] (\tikzinputsegmentfirst)
        .. controls
        (\tikzinputsegmentsupporta) and (\tikzinputsegmentsupportb)
        ..
        (\tikzinputsegmentlast);
      },
      closepath code={
        \path [#1]
        (\tikzinputsegmentfirst) -- (\tikzinputsegmentlast);
      },
    },
  },
  mid arrow/.style={postaction={decorate,decoration={
        markings,
        mark=at position .5 with {\arrow[#1]{stealth}}
      }}},
}

\graphicspath{{./figures/}}

\newcommand\copyrighttext{%
    \small \begin{center} \color{red} \textcopyright\,Accepted for presentation to the ``2nd Formal methods techniques in robotics systems: Design and control" Workshop at IROS'24, Abu Dhabi, UAE. Personal use of this material is permitted. Permission from authors must be obtained for all other uses, in any current or future media, including reprinting/republishing this material for advertising or promotional purposes, creating new collective works, for resale or redistribution to servers or lists, or reuse of any copyrighted component of this work in other works. \end{center}}
\newcommand\copyrightnotice{%
	\begin{tikzpicture}[remember picture,overlay]
	\node[anchor=south,yshift=25.6cm] at (current page.south) 
	{\color{red}\fbox{\parbox{\dimexpr\textwidth-\fboxsep-\fboxrule\relax}{\copyrighttext}}};
	\end{tikzpicture}%
}

\title{\copyrightnotice \LARGE \bf Task Coordination and Trajectory Optimization for Multi-Aerial Systems via Signal Temporal Logic: A Wind Turbine Inspection Study}

\author{Giuseppe Silano$^{1\star}$, Alvaro Caballero$^{2}$, Davide Liuzza$^{3,4}$, Luigi Iannelli$^{4}$, Stjepan Bogdan$^{5}$, and Martin Saska$^{1}$  
    \thanks{$^1$Department of Cybernetics, Czech Technical University in Prague, Prague, Czech Republic. 
    $^2$Department of System Engineering and Automation, University of Seville, Seville, Spain.
    $^3$Department of Engineering, University of Sannio in Benevento, Benevento, Italy.
    $^4$Fusion and Technology for Nuclear Safety and Security Department, ENEA, Frascati, Italy.
    $^5$Faculty of Electrical Engineering and Computing, University of Zagreb, Zagreb, Croatia.
    $^\star$Corresponding author: {\tt giuseppe.silano@fel.cvut.cz}.}
    \thanks{This publication is part of the R+D+i project TED2021-131716B-C22, funded by MCIN/AEI/10.13039/501100011033 and by the EU NextGenerationEU/PRTR. This work was also supported by the EU’s H2020 AERIAL-CORE grant no. 871479, by the ECSEL JU  COMP4DRONES grant no. 826610, by the Czech Science Foundation (GAČR) grant no. 23-07517S, by the CTU grant no. SGS23/177/OHK3/3T/13, and by the EU's project Robotics and advanced industrial production (reg. no. CZ.02.01.01/00/22 008/0004590).}
} 

\begin{document}

\maketitle
\thispagestyle{empty} 
\pagestyle{empty} 


\begin{acronym}
    \acro{CBF}[CBF]{Control Barrier Function}
    \acro{CVRP}[CVRP]{Capacity Vehicle Routing Problem}
    \acro{ILP}[ILP]{Integer Linear Programming}
    \acro{LP}[LP]{Linear Programming}
    \acro{LSE}[LSE]{Log-Sum-Exponential}
    \acro{LTL}[LTL]{Linear Temporal Logic}
    \acro{MILP}[MILP]{Mixed-Integer Linear Programming}
    \acro{MRS}[MRS]{Multi-Robot System}
    \acro{MTL}[MTL]{Metric Temporal Logic}
    \acro{NLP}[NLP]{Nonlinear Programming}
    \acro{PSO}[PSO]{Particle Swarm Optimization}
    \acro{ROS}[ROS]{Robot Operating System}
    \acro{SIL}[SIL]{Software-in-the-loop}
    \acro{STL}[STL]{Signal Temporal Logic}
    \acro{UAV}[UAV]{Unmanned Aerial Vehicle}
    \acro{TL}[TL]{Temporal Logic}
    \acro{TS}[TS]{Transition System}
    \acro{TSP}[TSP]{Traveling Salesman Problem}
    \acro{TWTL}[TWTL]{Time Window Temporal Logic}
    \acro{VRP}[VRP]{Vehicle Routing Problem}
    \acro{wSTL}[wSTL]{weighted-STL}
    \acro{wrt}[w.r.t.]{with respect to}
\end{acronym}



\begin{abstract}
    This paper presents a method for task allocation and trajectory generation in cooperative inspection missions using a fleet of multirotor drones, with a focus on wind turbine inspection. The approach generates safe, feasible flight paths that adhere to time-sensitive constraints and vehicle limitations by formulating an optimization problem based on \ac{STL} specifications. An event-triggered replanning mechanism addresses unexpected events and delays, while a generalized robustness scoring method incorporates user preferences and minimizes task conflicts. The approach is validated through simulations in MATLAB and Gazebo, as well as field experiments in a mock-up scenario.
\end{abstract}



\section*{Full-version}
\label{sec:fullVersion}

\begin{sloppypar}
The full-version~\cite{SilanoRAS2024} is available at~\href{https://doi.org/10.1016/j.robot.2024.104905}{10.1016/j.robot.2024.104905}.
\end{sloppypar}



\section{Introduction}
\label{sec:introduction}

\acp{UAV} are increasingly used for inspecting civilian infrastructure such as pipelines, wind turbine blades, power lines, towers, and bridges. Traditionally, these inspections rely on skilled operators and onboard sensors, but this approach carries safety risks, high costs, and potential human error. Although automated \ac{UAV} inspection methods have been proposed, they still face challenges related to navigation reliability, battery limitations, radio interference, and unpredictable environmental conditions \cite{Ollero2024BookChapter}.

This work presents a novel approach using \acf{STL} for task assignment and trajectory generation in collaborative multi-\ac{UAV} inspection missions. \ac{STL} is employed for its capacity to define complex mission objectives and time constraints in a natural language-like format while providing robustness metric to quantify how well these requirements are met. The approach formulates an optimization problem to generate safe, feasible trajectories that maximize robustness under dynamic constraints.

A wind turbine inspection serves as a case study to validate the method, which involves coordinating multiple \acp{UAV} to avoid collisions, manage time-bound tasks, and respect vehicle dynamics while ensuring full coverage of he infrastructure. A two-step hierarchical strategy is used: first, a \ac{MILP} model simplifies the initial mission objectives, and then a global \ac{STL} optimizer refines the trajectories based on the \ac{MILP} solution. The method also includes an event-triggered replanning mechanism to adjust \ac{UAV} trajectories in response to unforeseen events and a robustness scoring technique to address user preferences and task conflicts. Its effectiveness is demonstrated through MATLAB and Gazebo simulations and field experiments in a mock-up scenario.



\section{Problem Description}
\label{sec:problemDescription}

This paper focuses on coordinating a fleet of multirotor \acp{UAV} for collaborative inspection missions, using wind turbine inspection as a case study. The mission involves capturing videos and images of the turbine's pylon and blades, divided into two tasks: \textit{pylon inspection}, which evaluates the structural integrity of components like the nacelle and rotor shaft to detect hazards such as corrosion or damage, and \textit{blade inspection}, which scans the entire blade surface for cracks or coating damage, requiring precise \ac{UAV} control for full coverage without blur. 
The planning process must consider the turbine's size and \ac{UAV} dynamics, including heterogeneous velocity and acceleration limits, to generate efficient trajectories that maintain safe distances from obstacles and other \acp{UAV}. A pre-existing map with a polyhedral representation of obstacles is assumed for mission planning.



\section{Problem Solution}
\label{sec:problemSolution}

Consider a discrete-time dynamical model of a \ac{UAV} represented as $x_{k+1}=f(x_k, u_k)$, where $x_{k+1}, x_k \in \mathcal{X} \in \mathbb{R}^n$ are the system states, and $u_k \in \mathcal{U} \in \mathbb{R}^m$ is the control input. The time vector $\mathbf{t}=(t_0, \dots, t_N)^\top \in \mathbb{R}^{N+1}$, with $N$ samples and sampling period $T_s$. The state sequence $\mathbf{x}$ and control input sequence $\mathbf{u}$ for the $d$-th multirotor~\ac{UAV} are defined as ${^d}\mathbf{x}=({^d}\mathbf{p}^{(1)}, {^d}\mathbf{v}^{(1)}, {^d}\mathbf{p}^{(2)}, {^d}\mathbf{v}^{(2)}, {^d}\mathbf{p}^{(3)}, {^d}\mathbf{v}^{(3)})^\top$ and ${^d}\mathbf{u}=({^d}\mathbf{a}^{(1)}, {^d}\mathbf{a}^{(2)}, {^d}\mathbf{a}^{(3)})^\top$, where ${^d}\mathbf{p}^{(j)}$, ${^d}\mathbf{v}^{(j)}$, and ${^d}\mathbf{a}^{(j)}$ represent the \ac{UAV}'s position, velocity, and acceleration sequences along the $j$-axis of the world frame. The $k$-th elements of ${^d}\mathbf{p}^{(j)}$, ${^d}\mathbf{v}^{(j)}$, ${^d}\mathbf{a}^{(j)}$, and $\mathbf{t}$ are denoted as ${^d}p_k^{(j)}$, ${^d}v_k^{(j)}$, ${^d}a_k^{(j)}$, and $t_k$. 
The \ac{STL} grammar (omitted for brevity) is used to encode the inspection problem as:

\vspace*{-0.9em}
\small
\begin{equation}\label{eq:windTurbineInspectionFormula}
    \begin{split}
    \varphi =&  \bigwedge_{d\in\mathcal{D}}\square_{[0,T_N]} ( {^d}\varphi_{\mathrm{ws}}\wedge {^d}\varphi_{\mathrm{obs}}\wedge {^d}\varphi_{\mathrm{dis}} ) \,
    \wedge \\
    & \bigwedge_{q=1}^{N_\mathrm{tr}}\lozenge_{[0,T_N-T_{\mathrm{ins}}]} \bigvee_{d\in\mathcal{D}}\square_{[0,T_{\mathrm{ins}}]} {^d}\varphi_{\mathrm{tr,q}} \,
    \wedge \\
    & \bigwedge_{q=1}^{N_\mathrm{bla}}\lozenge_{[0,T_N-T_{\mathrm{bla}}]} \bigvee_{d\in\mathcal{D}}\square_{[0,T_{\mathrm{bla}}]} {^d}\varphi_{\mathrm{bla,q}} \, 
    \wedge \\
    & \bigwedge_{d\in\mathcal{D}} \lozenge_{[1,T_N]}{^d}\varphi_{\mathrm{hm}}\, 
    \wedge \\
    & \bigwedge_{d\in\mathcal{D}} \square_{[1,T_N-1]}\left( {^d}\varphi_{\mathrm{hm}} \hspace{-0.45em} \implies \hspace{-0.45em} \bigcirc_{[0, t_k + 1]} {^d}\varphi_{\mathrm{hm}} \right).
   \end{split}
\end{equation}

\normalsize
This \ac{STL} formula $\varphi$ encodes both safety and task requirements for the collaborative inspection mission of $\delta$ \acp{UAV} in the set $\mathcal{D}$ over the mission duration $T_N$. \textit{Safety requirements} include remaining in the workspace (${^d}\varphi_\mathrm{ws}$), avoiding obstacles (${^d}\varphi_\mathrm{obs}$), and maintaining safe distances between~\acp{UAV} (${^d}\varphi_\mathrm{dis}$). \textit{Task requirements} include pylon inspection (${^d}\varphi_\mathrm{tr}$), which involves visiting the $N_\mathrm{tr}$ target areas for at least $T_\mathrm{ins}$, and blade inspection (${^d}\varphi_\mathrm{bla}$), which requires scanning the blade surface at a specific distance and speed for at least $T_\mathrm{bla}$, with $N_\mathrm{bla}$ indicating the blade sides to cover. Each \ac{UAV} must also return to its initial position (${^d}\varphi_\mathrm{hm}$). The following equations define these specifications:

\vspace*{-0.9em}
\small
\begin{subequations}\label{eq:STLcomponents}
    \begin{align}
    \textstyle{ {^d}\varphi_\mathrm{ws}} &= \textstyle{ \bigwedge_{j=1}^3 {^d}\mathbf{p}^{(j)} \hspace{-0.25em} \in (\underline{p}^{(j)}_\mathrm{ws}, \bar{p}^{(j)}_\mathrm{ws})}, \label{subeq:belongWorkspace} \\
    \textstyle{ {^d}\varphi_\mathrm{obs}} &= \textstyle{ \bigwedge_{j=1}^3\bigwedge_{q=1}^{N_\mathrm{obs}} {^d}\mathbf{p}^{(j)} \hspace{-0.25em} \not\in (\underline{p}_{\mathrm{obs,q}}^{(j)}, \bar{p}_{\mathrm{obs,q}}^{(j)})}, \label{subeq:avoidObostacles} \\
    \textstyle{{^d}\varphi_\mathrm{hm}} &= \textstyle{\bigwedge_{j=1}^3 {^d}\mathbf{p}^{(j)} \hspace{-0.25em} \in (\underline{p}^{(j)}_\mathrm{hm}, \bar{p}^{(j)}_\mathrm{hm})}, \label{subeq:backHome} \\
    \textstyle{{^d}\varphi_\mathrm{dis}} &=  \textstyle{\bigwedge_{ \{d, m\} \in \mathcal{D}, d \neq m } \hspace{0.2em} \lVert {^d}\mathbf{p} - {^m}\mathbf{p} \rVert \geq \Gamma_\mathrm{dis}}, \label{subeq:keepDistance} \\
    %
    %
    \textstyle{{^d}\varphi_{\mathrm{tr,q}}} &=  \textstyle{\bigwedge_{j=1}^3  {^d}\mathbf{p}^{(j)} \hspace{-0.25em} \in (\underline{p}^{(j)}_{\mathrm{tr,q}}, \bar{p}^{(j)}_{\mathrm{tr,q}})}, \label{subeq:visitTargets} \\
    \textstyle{{^d}\varphi_\mathrm{bla,q}} &= \textstyle{\bigwedge_{j=1}^3 {^d}\mathbf{p}^{(j)} \hspace{-0.25em} \in (\underline{p}^{(j)}_{\mathrm{bla,q}}, \bar{p}^{(j)}_{\mathrm{bla,q}}) \; \wedge  \nonumber} \\ 
     & \hspace{1em} \textstyle{\mathrm{dist}_{\mathrm{bla,q}}({^d}\mathbf{p})\in (\Gamma_{\mathrm{bla}}-\varepsilon, \Gamma_{\mathrm{bla}}+\varepsilon)}. \label{subeq:bladeInspection} 
    \end{align}
\end{subequations} 

\normalsize
Here $\underline{p}^{(j)}_\bullet$ and $\bar{p}_\bullet^{(j)}$ define the rectangular boundaries for the workspace, obstacles, home positions, and target areas. The parameter $N_\mathrm{obs}$ represent the number of obstacles. $\Gamma_\mathrm{dis}$ is the minimum separation between \acp{UAV}, and $\mathrm{dist}_\mathrm{bla}(\cdot)$ calculates the Euclidean distance between a \ac{UAV} and the blade surface. The parameters $\Gamma_\mathrm{bla}$ and $\varepsilon$ represent the minimum required distance from the blade and the maneuverability margin, respectively.

To solve the problem, the robustness $\rho_\varphi(\mathbf{x}, t_k)$ is approximated with $\tilde{\rho}_\varphi(\mathbf{x}, t_k)$, the following optimization problem is formulated:

\vspace*{-0.9em}
\small
\begin{equation}\label{eq:optimizationProblemMotionPrimitives}
    \begin{split}
    &\maximize_{{^d}\mathbf{p}^{(j)}, \, {^d}\mathbf{v}^{(j)}, \,{^d}\mathbf{a}^{(j)}\atop d \in \mathcal{D}} \;\;
    {\tilde{\rho}_\varphi (\mathbf{p}^{(j)}, \mathbf{v}^{(j)} )} \\
    &\qquad \,\;\, \text{s.t.}~\quad\, {^d}\underline{v}^{(j)} \leq {^d}v^{(j)}_k \leq {^d}\bar{v}^{(j)}, \\
    &\,\;\;\;\, \qquad \qquad {^d}\underline{a}^{(j)} \leq {^d}a^{(j)}_k \leq {^d}\bar{a}^{(j)}, \\
    &\,\;\;\;\, \qquad \qquad \tilde{\rho}_\varphi ({^d}\mathbf{p}^{(j)}, \, {^d}\mathbf{v}^{(j)}) \geq \zeta, \\
    &\,\;\;\;\, \qquad \qquad {^d}\mathbf{S}^{(j)}, \forall k=\{0,1, \dots, N-1\}
    \end{split},
\end{equation}

\normalsize
where ${^d}\underline{v}^{(j)}$ and ${^d}\underline{a}^{(j)}$ represent the lower velocity and acceleration limits, and ${^d}\bar{v}^{(j)}$ and ${^d}\bar{a}^{(j)}$ denote their upper limits for drone $d$ along each $j$-axis of the world frame. The minimum robustness threshold $\tilde{\rho}_\varphi (\mathbf{p}^{(j)}, \mathbf{v}^{(j)}) \geq \zeta$, provides as a safety buffer to ensure the \ac{STL} formula $\varphi$ is satisfied even with disturbances. The shorthand ${^d}\mathbf{S}^{(j)}$ represents the motion primitives that follow the drone's dynamics along each axis. 

The resulting problem is a nonlinear, non-convex max-min optimization, which is challenging to solve in a reasonable time due to solvers' tendency to converge to local optima. To address this, we compute an initial guess using a simplified \ac{MILP} formulation based on a subset of the original \ac{STL} specification $\varphi$:

\vspace*{-1em}
\small
\begin{subequations}\label{eq:MILP}
    \begin{align}
        &\minimize_{z_{ij|d}}
        { \sum\limits_{ \{i,j \} \in \mathcal{V}, \, i \neq j, \, d \in \mathcal{D}} \hspace{-0.30cm} w_{ij|d} \, z_{ij|d} } \label{subeq:objectiveFunction} \\
        %
        &\quad \text{s.t.} \hspace{0cm} \sum\limits_{i \in \mathcal{N}_j^\mathrm{in}} \hspace{-0.05cm} z_{ij|d} = \hspace{-0.25cm} \sum\limits_{i \in \mathcal{N}_j^\mathrm{out}} \hspace{-0.05cm} z_{ji|d}, \; \forall j \in \mathcal{T}, \; \forall d \in \mathcal{D}, \label{subeq:notAccumulating} \\ 
        &\qquad \, \; \sum\limits_{ d \in \mathcal{D}} \, \sum\limits_{ i \in \mathcal{N}_j^\mathrm{in}} \quad \hspace{-0.15cm} z_{ij|d} \geq 1, \; \forall j \in \mathcal{T}, \label{subeq:visitedOneUAV} \\ 
        &\qquad \hspace{-0.04cm} \sum\limits_{ j \in \mathcal{N}_{o_d}^\mathrm{out} } \hspace{-0.175cm} z_{o_d j|d} = 1, \; \forall d \in \mathcal{D}, \label{subeq:arrivalDepartureDepot} \\ 
        &\qquad \hspace{0.025cm} \sum\limits_{j \in \mathcal{N}_{o_d}^\mathrm{in} } \hspace{-0.175cm} z_{jo_d|d} = 1, \; \forall d \in \mathcal{D}. \label{subeq:arrivalDepartureSameDepot} 
     \end{align}
\end{subequations}

\normalsize
The objective function~\eqref{subeq:objectiveFunction} minimizes the total flight time for the fleet of~\acp{UAV}, while ensuring that drones do not revisit the same edges unnecessarily. The constraints ensure that \acp{UAV} do not accumulate in an area \eqref{subeq:notAccumulating}, that all target areas and blade points are visited at least once \eqref{subeq:visitedOneUAV}, and that each \ac{UAV} starts and returns to its depot \eqref{subeq:arrivalDepartureDepot} and \eqref{subeq:arrivalDepartureSameDepot}. Any disconnected subtours in the \ac{MILP} solution are connected afterwards, as the \ac{MILP} primarily serves to seed the final \ac{STL} optimizer. This allows us to save computational time without enforcing complex subtour elimination constraints. $\mathcal{N}_i^\mathrm{in}$ denotes the \textit{in-neighborhood}, the set of nodes with an edge entering $i$, i.e., $\mathcal{N}_i^\mathrm{in}=\{j\in \mathcal{V}: (j,i)\in\mathcal{E}\}$. Similarly, \textit{out-neighborhood} is the set of nodes with an entering edge which starts from $i$, i.e., $\mathcal{N}_i^\mathrm{out}=\{j\in \mathcal{V}: (i,j)\in\mathcal{E}\}$.

Due to space constraints, details on the event-triggered replanning mechanism and the generalized robustness scoring method, which dynamically adjust \ac{UAV} trajectories and resolve conflicts between competing objectives, are omitted. For more information, see \cite{SilanoRAS2024}.



\section{Experimental Results}
\label{sec:experimentalResults}

The effectiveness and validity of the approach are demonstrated through simulations in MATLAB and Gazebo, as well as field experiments carried out in a mock-up scenario. Videos that can be accessed at~\url{https://mrs.fel.cvut.cz/milp-stl}.



\balance
\bibliographystyle{IEEEtran}
\bibliography{bib_short}

\end{document}